\documentclass[letterpaper,10pt,conference]{ieeeconf}

\IEEEoverridecommandlockouts
\usepackage{graphicx}
\usepackage{array}
\usepackage{xcolor}
\usepackage{url}
\usepackage{amsmath}
\usepackage{amssymb}
\usepackage{booktabs}
\usepackage{multirow}
\usepackage{subcaption}

\title{\LARGE \bf {RevoSim: Scalable Multimodal Tactile Simulation for Dexterous Manipulation}}

\author{\authorblockN{Ke Zhang$^{1,2*}$, Lixin Xu$^{1,2*}$, Ziyi Wang$^{3*}$, Xiyue Dong$^{1,2*}$,\\
Jie Tan$^{2}$, Lianghu Shao$^{2}$, Xianpeng Wang$^{1,2}$,
Chuanyu Li$^{2}$, Renjing Xu$^{1\dagger}$ \\
${}^{1}$HKUST(GZ), ${}^{2}$BrainCo, ${}^{3}$Georgia Tech, \\
$*$ Equal Contribution  $\dagger$ Corresponding Author \\
\url{https://brainco.github.io/RevoSim-web/}
}}

\begin{document}

\maketitle
\thispagestyle{empty}
\pagestyle{empty}

\begin{abstract}
We present RevoSim, a scalable and high-fidelity multimodal tactile simulator
for the BrainCo Revo3 hand that jointly models 285 piezoresistive taxels and
five visuotactile fingertips. RevoSim efficiently generates synchronized
pressure, depth, RGB, and marker-displacement observations; on a single RTX
5090, its depth-conditioned marker-displacement configuration supports up to
4,096 parallel environments. We train in-hand manipulation and grasping
policies entirely in simulation, establishing a foundation for future
sim-to-real transfer.
\end{abstract}

\section{Introduction}
\label{sec:introduction}

Dexterous manipulation is fundamentally contact-rich. Vision and proprioception provide estimates of object and robot states, but cannot directly observe local interaction phenomena such as incipient slip, contact migration, or hidden contact. Tactile feedback is therefore crucial once contact occurs. However, scaling tactile reinforcement learning remains challenging: physical hardware wears during data collection, tactile observations are high-dimensional and device-specific, and fast, transferable simulation is still unavailable for many tactile embodiments~\cite{chung2026tactilegenesis,akinola2024tacsl}.

Existing tactile simulators can synthesize fingertip images and marker motion,
model compliant deformation, or provide efficient tactile approximations for
parallel reinforcement learning~\cite{wang2022tacto,si2022taxim,zhao2024fots,
du2024tacipc,si2024difftactile,akinola2024tacsl,nguyen2024tacex}.
Distributed-array simulators further support force fields, configurable taxel
layouts, and sensor non-idealities~\cite{akinola2024tacsl,
chung2026tactilegenesis}. However, most methods focus on a single sensor family
or generic taxel layout. Accurately reproducing heterogeneous whole-hand
tactile sensing while maintaining the throughput required for parallel
reinforcement learning therefore remains challenging.

To address this gap, we present \emph{RevoSim}, a scalable multimodal tactile
simulator for the \textbf{BrainCo Revo3} hand~\cite{brainco2026revo3}. Revo3
combines vision-based tactile fingertips, which capture local contact geometry
and elastomer motion, with piezoresistive arrays that measure contact locations
and loads across the palm and finger pads. To the best of our knowledge,
RevoSim is the first framework to jointly simulate these two modalities,
generating synchronized high-resolution fingertip observations and whole-hand
taxel measurements in parallel environments, as shown in
Fig.~\ref{fig:tactile_visualization}. Both modalities are calibrated against real
sensor responses, including their geometry, sensitivity, temporal behavior,
drift, hysteresis, dead taxels, and crosstalk. Qualitative comparisons show
consistency between simulated and real sensor responses. The simulator
supports policy training in simulation and provides a basis for future
sim-to-real transfer studies.

\section{Method}
\label{sec:method}

\subsection{System Overview}

\begin{figure}[t]
    \centering

    \begin{subfigure}[t]{0.41\linewidth}
        \centering
        \includegraphics[width=\linewidth]{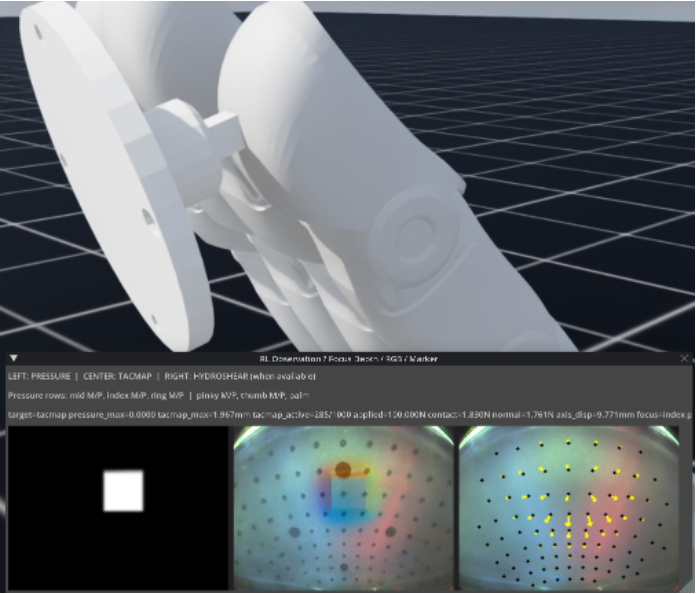}
        \caption{Visuo-tactile simulation with depth, RGB, and markers.}
        \label{fig:result_a}
    \end{subfigure}
    \hfill
    \begin{subfigure}[t]{0.57\linewidth}
        \centering
        \includegraphics[width=\linewidth]{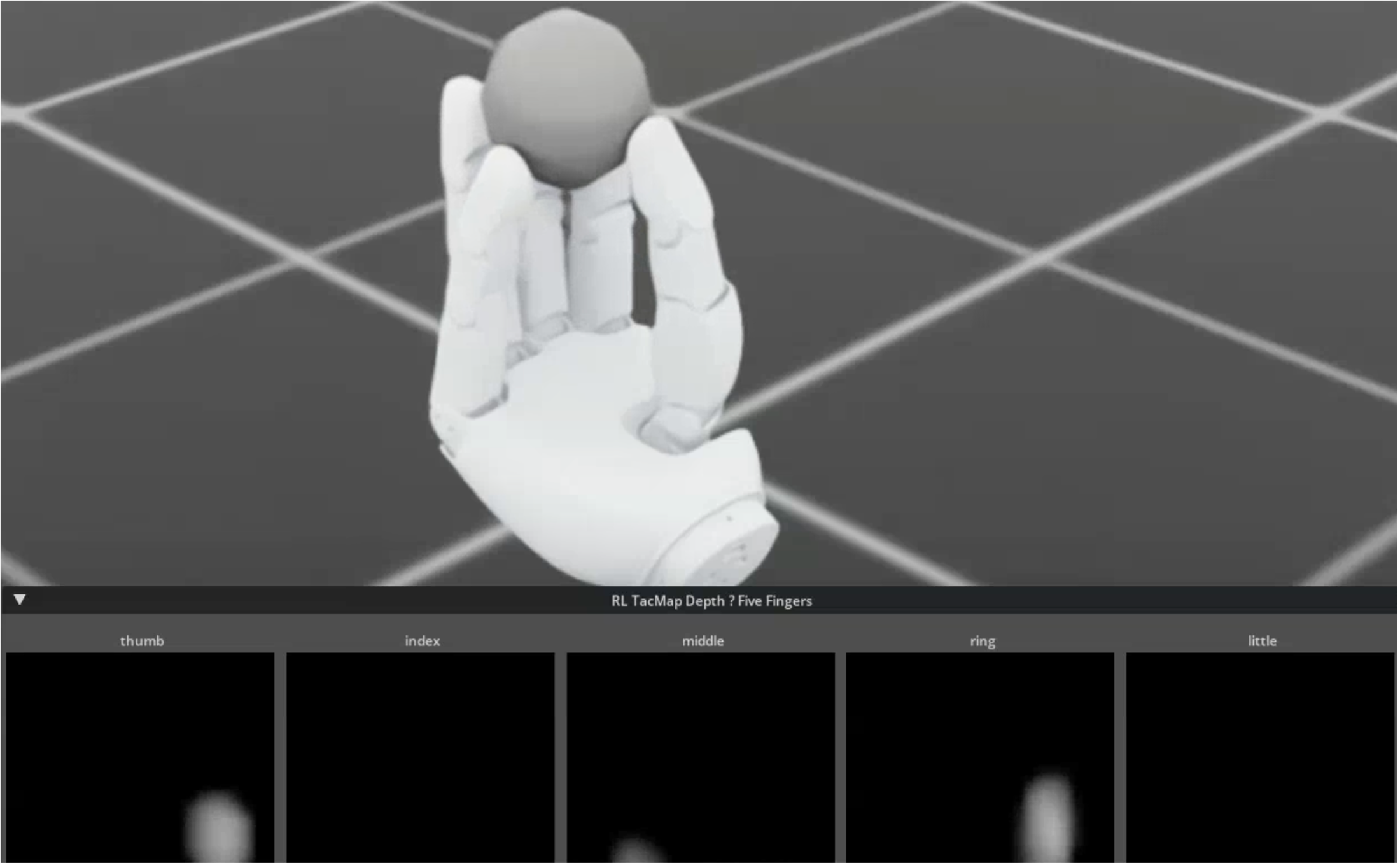}
        \caption{In-hand ball rotation trained with visuo-tactile depth observation.}
        \label{fig:result_b}
    \end{subfigure}

    \vspace{2mm}

    \begin{subfigure}[t]{\linewidth}
        \centering
        \includegraphics[width=\linewidth]{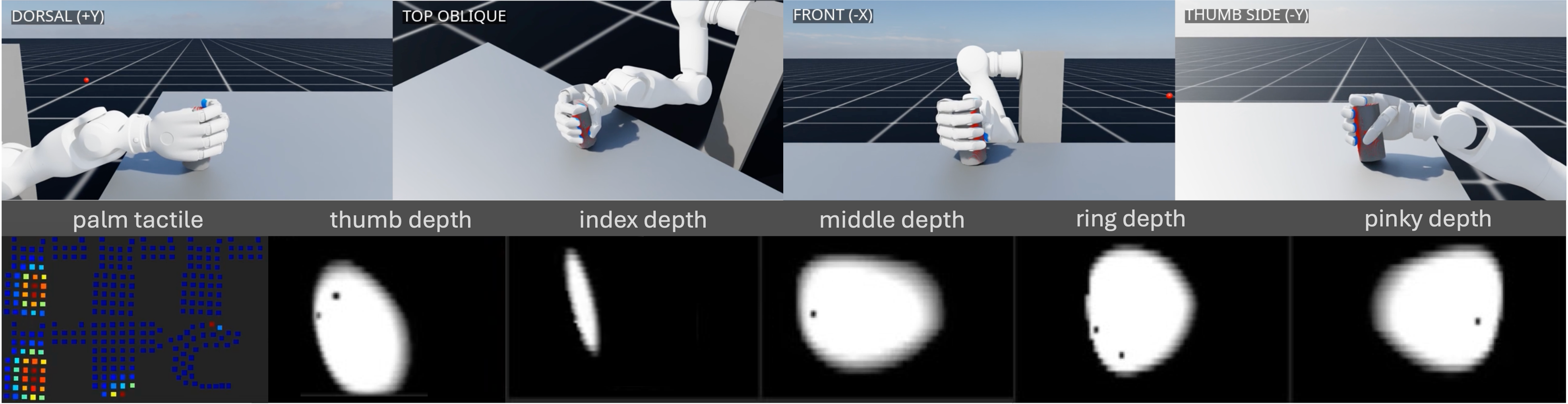}
        \caption{Object grasping trained with full-palm tactile and fingertip
        visuo-tactile observation.}
        \label{fig:result_c}
    \end{subfigure}

    \caption{RevoSim observations and tactile-learning tasks.}
    \label{fig:tactile_visualization}
\end{figure}

RevoSim models the Revo3 hand's eleven pressure regions and five visuotactile
fingertips. At each step, it produces
\begin{equation}
\mathcal{O}_t=\{\mathbf{p}_t,\mathbf{D}_t,\mathbf{I}_t,\mathbf{M}_t\},
\end{equation}
where $\mathbf{p}_t\in\mathbb{R}_{+}^{285}$ is the whole-hand pressure vector,
$\mathbf{D}_t\in\mathbb{R}_{+}^{5\times240\times320}$ and
$\mathbf{I}_t\in[0,1]^{5\times3\times240\times320}$ are fingertip depth
and RGB images, and $\mathbf{M}_t\in\mathbb{R}^{5\times100\times3}$ contains
the 3D displacements of 100 calibrated markers on each fingertip. Image-space
marker flow is derived from $\mathbf{M}_t$ only for rendering.
Fig.~\ref{fig:result_a} illustrates the fingertip outputs.
Ball rotation uses five-finger depth (Fig.~\ref{fig:result_b}), while grasping
combines palm pressure with fingertip depth (Fig.~\ref{fig:result_c}).

\subsection{Geometry-Aware Pressure Simulation}

For taxel $i$, let $\mathbf{x}_{i,t}$ and unit $\mathbf{n}_{i,t}$ denote its
world-frame position and outward normal, and let $\phi_t$ be the object's signed
distance function, negative inside the object. The code computes penetration
and penetration rate as
\begin{equation}
\delta_{i,t}=[-\phi_t(\mathbf{x}_{i,t})]_+,
\qquad
\dot{\delta}_{i,t}=\frac{\delta_{i,t}-\delta_{i,t-1}}{\Delta t}.
\label{eq:pressure-penetration}
\end{equation}
With per-taxel calibration parameters stiffness $k_i$, damping $c_i$, gain
$g_i$, bias $b_i$, threshold $\theta_i$, saturation force $F_i$, and response
exponent $\gamma_i$, the undiffused pressure is
\begin{equation}
\begin{aligned}
r_{i,t}
&=\mathbf{1}_{\delta_{i,t}>0}
\left[g_i\left(k_i\delta_{i,t}+c_i\dot{\delta}_{i,t}\right)+b_i\right],\\
p_{i,t}
&=\left(
\frac{\operatorname{clip}\!\left(
\mathbf{1}_{r_{i,t}\geq\theta_i}r_{i,t},0,F_i\right)}{F_i}
\right)^{\gamma_i}.
\label{eq:pressure-calibration}
\end{aligned}
\end{equation}

To propagate a taxel response over the curved sensing surface without leaking
through oppositely oriented parts, define the tangent projector
$\mathbf{P}_i=\mathbf{I}-\mathbf{n}_i\mathbf{n}_i^\top$ (suppressing $t$) and the symmetric
tangent-plane distance
\begin{equation}
d_{ij}^2=\frac{1}{2}\left(
\|\mathbf{P}_i(\mathbf{x}_i-\mathbf{x}_j)\|_2^2+
\|\mathbf{P}_j(\mathbf{x}_i-\mathbf{x}_j)\|_2^2
\right).
\label{eq:pressure-distance}
\end{equation}
The unnormalized and normalized diffusion kernels are
\begin{equation}
\begin{aligned}
W_{ij}
&=\mathbf{1}_{d_{ij}\leq r_\sigma}
\exp\!\left(-\frac{d_{ij}^2}{2\sigma^2}\right)
\left[\max(\mathbf{n}_i^\top\mathbf{n}_j,0)\right]^q,\\
K_{ij}&=\frac{W_{ij}}{\sum_{a}W_{aj}},
\qquad
\widetilde{\mathbf{p}}_t
=(1-\rho)\mathbf{p}_t+\rho\mathbf{K}\mathbf{p}_t.
\label{eq:pressure-diffusion}
\end{aligned}
\end{equation}
Here $K_{ij}$ is the fraction of source taxel $j$ assigned to target taxel
$i$. Consequently, $\sum_i K_{ij}=1$ and the diffusion preserves the total
response on irregular layouts. We use $\sigma=3\,\mathrm{mm}$,
$r_\sigma=3\sigma$, $q=1$, and $\rho=0.7$.

\subsection{Adaptive Visuotactile Simulation}

Each fingertip has a calibrated $240\times320$ reference ray grid. For pixel
$a=(u,v)$, let $\mathbf{r}_a(\ell)=\mathbf{o}_a+\ell\mathbf{d}_a$ be its ray,
$\ell_a^{S}$ the cached intersection distance to the undeformed sensor surface,
and $v_a^{S}$ its validity. At every simulation step, a coarse $32\times24$
penetration map $D_t^c$ localizes contact. If $\boldsymbol{\xi}_a\in\mathbb{R}^2$
is the ray origin expressed in metric camera-plane coordinates, the contact
set and expanded region of interest (ROI) are
\begin{equation}
\mathcal{C}_t=\{\boldsymbol{\xi}_a\mid D_t^c(a)>\tau\},
\qquad
\Omega_t=\operatorname{BBox}(\mathcal{C}_t)\oplus[-m,m]^2.
\label{eq:adaptive-roi}
\end{equation}
Let $\chi_t=\mathbf{1}_{\mathcal{C}_t\neq\varnothing}$. If no coarse pixel is
active, the ROI falls back to the full calibrated camera plane. A centered
constant-integer-stride $25\times40$ lattice
$\mathcal{S}_t\subset\Omega_t$ is then indexed from the reference grid. Only
these rays are cast against the object. With object-hit distance
$\ell_{a,t}^{O}$ and validity $v_{a,t}^{O}$, the local indentation is
\begin{equation}
\begin{aligned}
\nu_{a,t}&=\mathbf{1}_{v_a^{S}\wedge v_{a,t}^{O}}
\mathbf{1}_{\ell_a^{S}-\ell_{a,t}^{O}>\varepsilon_D},\\
D_t^{\mathrm{loc}}(a)
&=\chi_t\nu_{a,t}\left[\ell_a^{S}-\ell_{a,t}^{O}\right]_+,
\qquad a\in\mathcal{S}_t.
\end{aligned}
\label{eq:adaptive-depth}
\end{equation}
The dense observation is reconstructed by bilinearly sampling the local patch
inside the ROI and masking invalid reference pixels,
\begin{equation}
\mathbf{D}_t=\mathcal{U}_{\Omega_t}
\!\left(D_t^{\mathrm{loc}};\,v^{S}\right).
\label{eq:dense-depth}
\end{equation}
Thus, a fixed ray budget follows contact while all output modalities remain on
the same calibrated $240\times320$ grid. The implementation uses
$\tau=0.02\,\mathrm{mm}$, $m=1\,\mathrm{mm}$, and
$\varepsilon_D=1\,\mu\mathrm{m}$.

\begin{figure*}[t]
    \centering
    \begin{subfigure}[t]{0.325\textwidth}
        \centering
        \includegraphics[width=\linewidth]{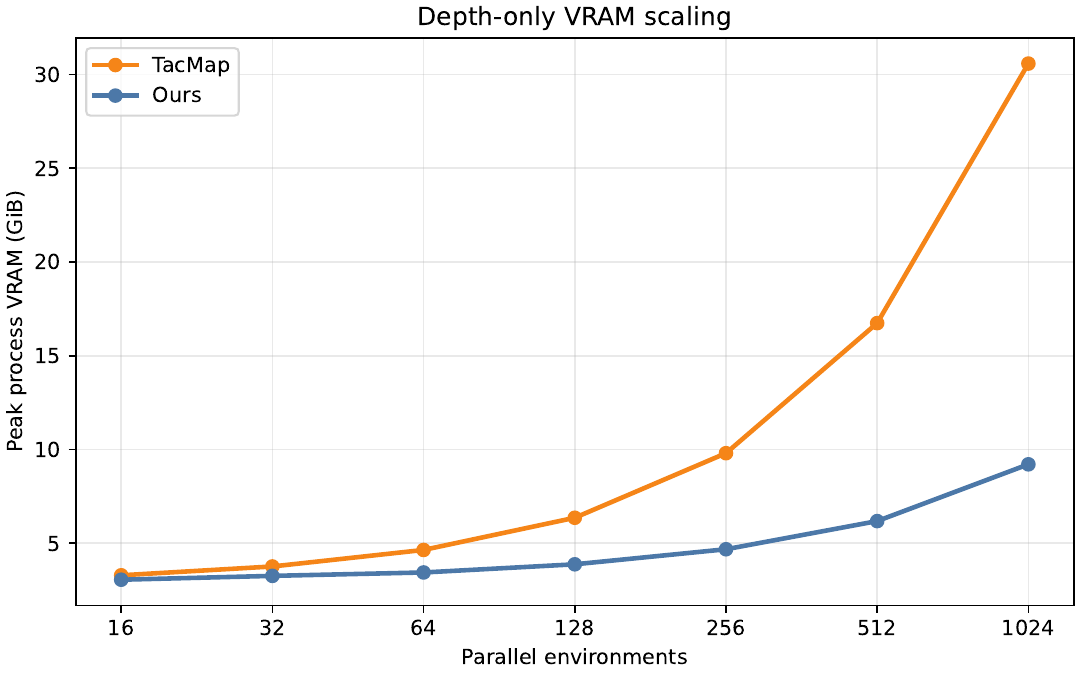}
        \caption{Depth only.}
        \label{fig:depth_vram_scaling}
    \end{subfigure}
    \hfill
    \begin{subfigure}[t]{0.325\textwidth}
        \centering
        \includegraphics[width=\linewidth]{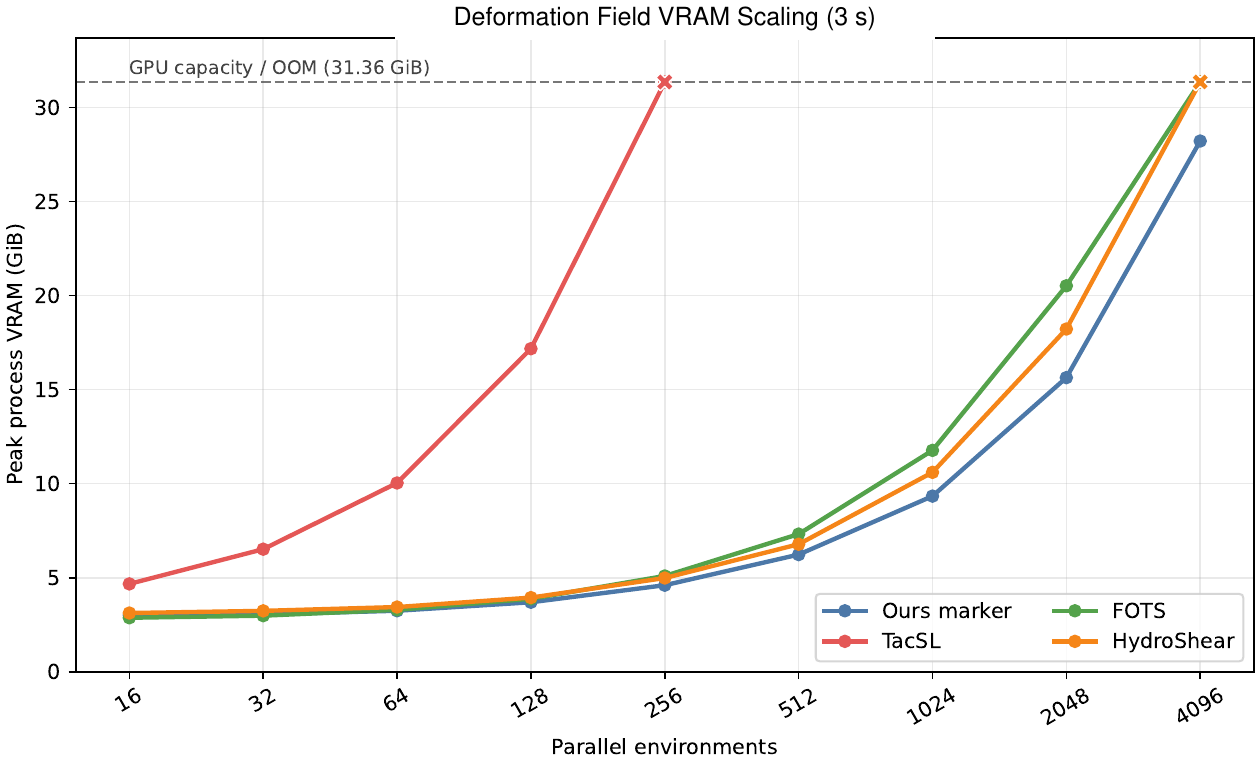}
        \caption{Deformation field.}
        \label{fig:tactile_field_vram_scaling}
    \end{subfigure}
    \hfill
    \begin{subfigure}[t]{0.325\textwidth}
        \centering
        \includegraphics[width=\linewidth]{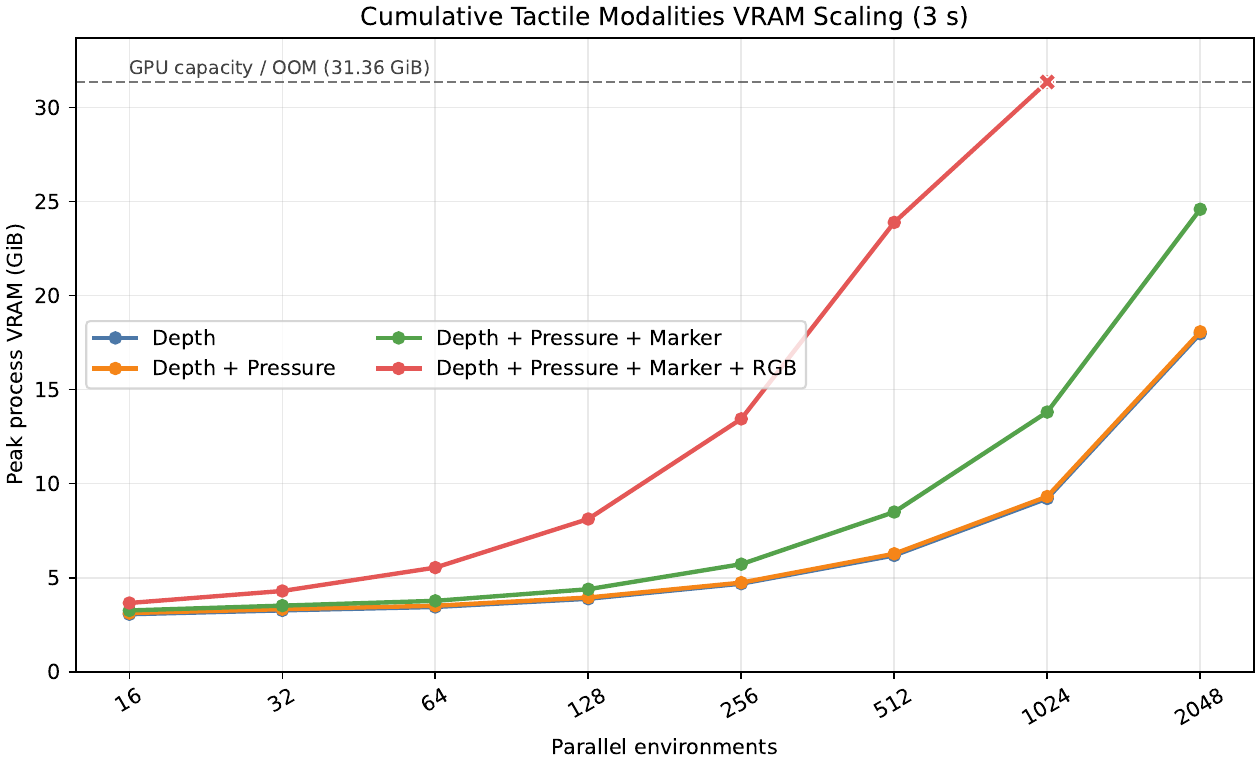}
        \caption{Combined modalities.}
        \label{fig:observation_vram_ablation}
    \end{subfigure}
    \caption{GPU-memory scalability of RevoSim.}
    \label{fig:vram_scaling}
\end{figure*}

\subsection{RGB and Marker Synthesis}

The dense indentation drives GPU-Taxim after conversion to its signed
millimeter height convention, $\mathbf{H}_t=-10^3\mathbf{D}_t$. Let
$\mathcal{T}_{\mathrm{Taxim}}$ denote the renderer. A real no-contact image is
combined with the simulated contact residual as
\begin{equation}
\begin{aligned}
\overline{\mathbf{I}}_t
&=\operatorname{clip}\!\left(
\mathbf{I}_{\mathrm{bg}}+\mathcal{T}_{\mathrm{Taxim}}(\mathbf{H}_t)
-\mathcal{T}_{\mathrm{Taxim}}(\mathbf{0}),0,255\right),\\
\mathbf{I}_t&=\overline{\mathbf{I}}_t/255.
\end{aligned}
\label{eq:taxim-rgb}
\end{equation}
and an inactive ROI returns $\mathbf{I}_{\mathrm{bg}}/255$ directly.

RevoSim models a spatial deformation field and samples it at the 100 calibrated
marker locations. These discrete samples form the marker-displacement
observation $\mathbf{M}_t$; marker displacement is therefore a policy input,
not a separate field model. Let sampled depth point $j$ have indentation
$h_{j,t}$ and calibrated camera-grid coordinate $\mathbf{u}_j$. With the camera-grid metric basis
$\mathbf{J}=[\mathbf{g}_u\ \mathbf{g}_v]\in\mathbb{R}^{3\times2}$, define
\begin{equation}
\mathbf{r}_{ij}=\mathbf{J}(\mathbf{u}_i-\mathbf{u}_j),
\qquad
d_{ij}^2=\|\mathbf{r}_{ij}\|_2^2.
\label{eq:marker-metric}
\end{equation}
The dilation field is evaluated as a surface quadrature over the adaptive
lattice,
\begin{equation}
\Delta\mathbf{x}_{i,t}^{\mathrm{dil}}
=a_d\sum_{j\in\mathcal{S}_t}
\omega_j h_{j,t}\,\mathbf{r}_{ij}
\exp(-\lambda_d d_{ij}^2),
\label{eq:marker-dilation}
\end{equation}
where $\omega_j$ is the sampled cell area divided by the calibrated marker-cell
area. This factor makes the result insensitive to the adaptive lattice stride.

Shear is temporal. For object sample $j$ at $\mathbf{o}_{j,t}$ in the
motion-compensated elastomer frame, let
$\varphi_{j,t}$ be its signed distance to the elastomer. The in-contact part of
its motion is
\begin{equation}
\begin{aligned}
\alpha_{j,t}&=
\frac{[-\varphi_{j,t}]_+-[-\varphi_{j,t-1}]_+}
{\varphi_{j,t}-\varphi_{j,t-1}},\\
\Delta\mathbf{q}_{j,t}&=\alpha_{j,t}
(\mathbf{o}_{j,t}-\mathbf{o}_{j,t-1}).
\end{aligned}
\label{eq:marker-contact-motion}
\end{equation}
with the finite fallback
$\alpha_{j,t}=\tfrac{1}{2}[1-\operatorname{sign}(\varphi_{j,t})]$ when the
ratio is undefined. It is decomposed using the current contact normal
$\mathbf{n}_{j,t}$:
\begin{equation}
\Delta q^n_{j,t}=\mathbf{n}_{j,t}^{\top}\Delta\mathbf{q}_{j,t},
\qquad
\Delta\mathbf{q}^{\tau}_{j,t}=
(\mathbf{I}-\mathbf{n}_{j,t}\mathbf{n}_{j,t}^{\top})
\Delta\mathbf{q}_{j,t}.
\label{eq:marker-motion-split}
\end{equation}
After transporting the previous tangential state into the current tangent
plane, the hydroelastic state update is
\begin{equation}
\begin{aligned}
\widehat f^n_{j,t}
&=\left[f^n_{j,t-1}+EA_j\Delta q^n_{j,t}\right]_+,\\
\widehat{\mathbf{f}}^{\tau}_{j,t}
&=\mathbf{P}_{j,t}\mathbf{f}^{\tau}_{j,t-1}
+KA_j\Delta\mathbf{q}^{\tau}_{j,t},\\
\mathbf{f}^{\tau}_{j,t}
&=\min\!\left(1,
\frac{\mu\widehat f^n_{j,t}}
{\|\widehat{\mathbf{f}}^{\tau}_{j,t}\|_2+\epsilon}\right)
\widehat{\mathbf{f}}^{\tau}_{j,t},
\end{aligned}
\label{eq:hydroelastic-state}
\end{equation}
where $\mathbf{P}_{j,t}=\mathbf{I}-\mathbf{n}_{j,t}\mathbf{n}_{j,t}^{\top}$.
The force state is reset to zero when sample $j$ leaves contact. The
tangential state shifts the Gaussian source in camera-grid coordinates,
\begin{equation}
\widehat{\mathbf{u}}_{j,t}=\mathbf{u}_{j,t}
+\mathbf{J}^{+}\mathbf{f}^{\tau}_{j,t},
\end{equation}
where $\mathbf{J}^{+}$ is the pseudoinverse of the grid metric. Let
$\boldsymbol{\Psi}(\mathbf{u})\in\mathbb{R}^3$ interpolate a UV coordinate on
the cached curved surface, and let $\mathbf{z}_i$ be marker $i$'s calibrated 3D
position. The source vector is transported to marker $i$'s tangent plane by
$\mathbf{P}_i=\mathbf{I}-\mathbf{n}_i\mathbf{n}_i^\top$. The resulting shear
displacement is
\begin{equation}
\Delta\mathbf{x}_{i,t}^{\mathrm{shr}}
=-a_s\sum_{j\in\mathcal{K}_t}
\widehat f^n_{j,t}\,\mathbf{P}_i\mathbf{f}^{\tau}_{j,t}
\exp\!\left[-\lambda_s
\|\mathbf{z}_i-\boldsymbol{\Psi}(\widehat{\mathbf{u}}_{j,t})\|_2^2\right],
\label{eq:marker-shear}
\end{equation}
where $\mathcal{K}_t$ contains active object samples. Finally,
\begin{equation}
\mathbf{M}_{i,t}=\Delta\mathbf{x}_{i,t}^{\mathrm{dil}}
+\Delta\mathbf{x}_{i,t}^{\mathrm{shr}}\in\mathbb{R}^{3}.
\label{eq:marker-total}
\end{equation}
For visualization only, the fingertip's calibrated image-plane axes
$\widehat{\mathbf{g}}_u,\widehat{\mathbf{g}}_v$ map this 3D observation to
pixel motion:
\begin{equation}
\begin{bmatrix}\Delta u_i\\ \Delta v_i\end{bmatrix}
=s_{\mathrm{px}}
\begin{bmatrix}
\widehat{\mathbf{g}}_u^\top\\
\widehat{\mathbf{g}}_v^\top
\end{bmatrix}\mathbf{M}_{i,t}.
\label{eq:marker-projection}
\end{equation}

\section{Experiments}
\label{sec:experiment}

\subsection{Setup and Metrics}

We implement RevoSim in Isaac Lab 2.3.2 on Ubuntu 24.04 with an RTX 5090 and
compare against TacMap~\cite{su2026tacmap}, TacSL~\cite{akinola2024tacsl},
FOTS~\cite{zhao2024fots}, and HydroShear~\cite{dang2026hydroshear} where their
modalities apply. We report peak per-process VRAM and maximum parallel
environments. Fig.~\ref{fig:vram_scaling} separates depth generation,
deformation-field simulation, and the cumulative cost of additional
observations. The latter two measurements cover three seconds of simulation;
crosses indicate out-of-memory configurations.

Qualitative sensor comparisons examine fingertip appearance and marker motion
(Fig.~\ref{fig:visuotactile_fidelity}), contact setups
(Fig.~\ref{fig:piezoresistive_contact}), and taxel responses
(Fig.~\ref{fig:piezoresistive_response}). Fig.~\ref{fig:result_b}
and Fig.~\ref{fig:result_c} illustrate their use in manipulation tasks.

\subsection{Computational Scalability}

\textbf{Depth and Deformation Fields.}
Fig.~\ref{fig:depth_vram_scaling} shows that RevoSim uses 9.2~GiB at 1,024
depth-only environments, approximately 70\% less than TacMap's 30.6~GiB.
The adaptive sampler maintains a fixed local ray budget while reconstructing
the calibrated dense observation. This separates the resolution of the
returned depth image from the number of fine contact queries.

Fig.~\ref{fig:tactile_field_vram_scaling} evaluates deformation-field
simulation; for RevoSim, the field is sampled as the depth-conditioned
marker-displacement observation $\mathbf{M}_t$. This configuration supports
4,096 environments within 31.36~GiB. FOTS and HydroShear run out of memory at
this scale, while TacSL reaches the memory limit at 256 environments.
RGB is excluded from this setting.

\textbf{Observation Ablation.}
Fig.~\ref{fig:observation_vram_ablation} adds pressure, marker motion, and
RGB to depth in sequence. Pressure adds little overhead, while marker motion
and RGB increase memory use; the full setting reaches capacity at 1,024
environments. These results motivate selecting observations for each task:
whole-hand pressure provides distributed contact information at a small
additional memory cost, while fingertip RGB and markers expose appearance
and deformation when needed. The plots measure memory capacity rather than
simulation throughput.

\subsection{Qualitative Sensor Comparisons}

\begin{figure}[t]
    \centering
    \captionsetup[subfigure]{justification=centering}
    \begin{subfigure}[t]{0.32\linewidth}
        \centering
        \includegraphics[width=\linewidth]{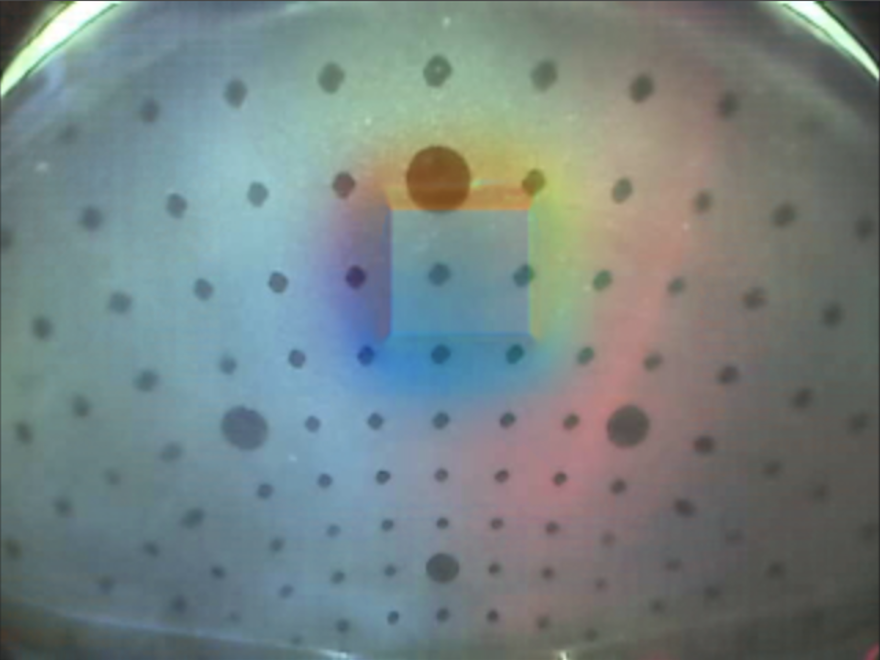}
        \caption{Simulated RGB.}
        \label{fig:simulated_rgb}
    \end{subfigure}
    \hfill
    \begin{subfigure}[t]{0.32\linewidth}
        \centering
        \includegraphics[width=\linewidth]{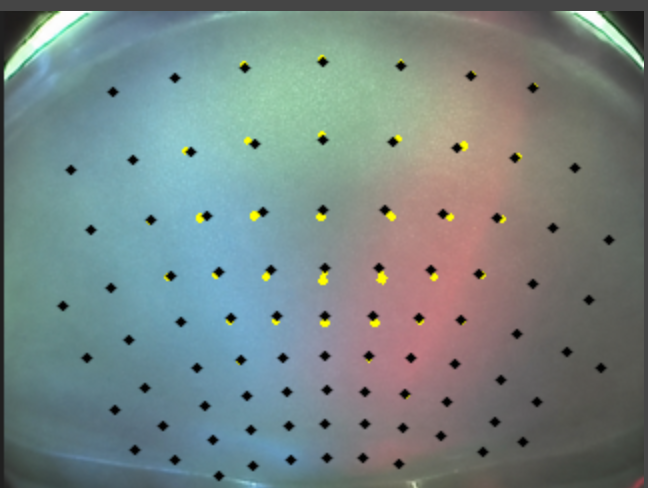}
        \caption{Simulated markers.}
        \label{fig:simulated_markers}
    \end{subfigure}
    \hfill
    \begin{subfigure}[t]{0.32\linewidth}
        \centering
        \includegraphics[width=\linewidth]{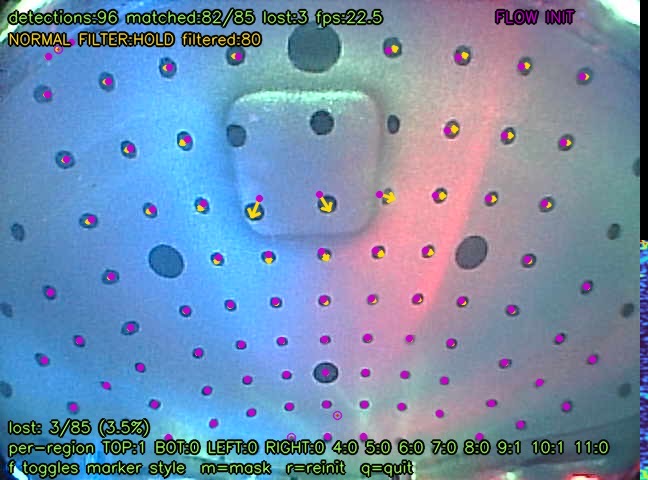}
        \caption{Real response.}
        \label{fig:real_visuotactile}
    \end{subfigure}
    \caption{Qualitative visuo-tactile comparison of simulated appearance
    and marker motion with a real sensor response.}
    \label{fig:visuotactile_fidelity}
\end{figure}

\textbf{Visuo-tactile Responses.}
Fig.~\ref{fig:visuotactile_fidelity} presents the optical and deformation
outputs separately. The simulated RGB image in Fig.~\ref{fig:simulated_rgb}
shows a localized indentation against the sensor's background illumination.
Fig.~\ref{fig:simulated_markers} visualizes marker displacement, while
Fig.~\ref{fig:real_visuotactile} shows the physical response with tracked
markers. Together, the images illustrate contact-induced appearance changes
and spatially varying marker motion. The background-residual construction
in Eq.~\eqref{eq:taxim-rgb} preserves the no-contact appearance, and the
deformation model supplies marker motion in calibrated coordinates.

\begin{figure}[t]
    \centering
    \captionsetup[subfigure]{justification=centering}
    \begin{subfigure}[t]{0.48\linewidth}
        \centering
        \includegraphics[width=\linewidth]{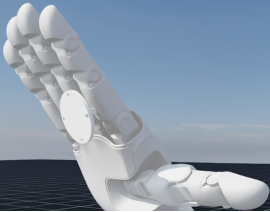}
        \caption{Simulated contact.}
        \label{fig:simulated_pressure_contact}
    \end{subfigure}
    \hfill
    \begin{subfigure}[t]{0.48\linewidth}
        \centering
        \includegraphics[width=\linewidth]{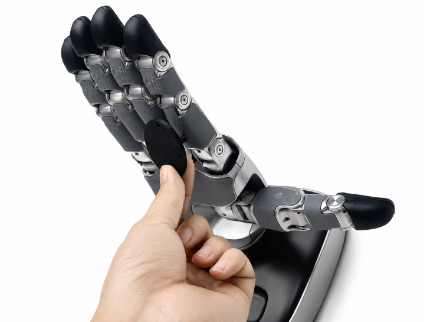}
        \caption{Physical pressing test.}
        \label{fig:real_pressure_contact}
    \end{subfigure}
    \caption{Simulated and physical contact setups for piezoresistive sensing.}
    \label{fig:piezoresistive_contact}
\end{figure}

\textbf{Contact Setups and Pressure Responses.}
Fig.~\ref{fig:piezoresistive_contact} shows the simulated contact geometry
and a physical pressing test. In simulation, object penetration at the taxel
locations drives the calibrated response, followed by diffusion over the
curved sensing surface. The normal-dependent kernel in
Eq.~\eqref{eq:pressure-diffusion} controls how that response spreads to
neighboring taxels.

\begin{figure}[t]
    \centering
    \captionsetup[subfigure]{justification=centering}
    \begin{subfigure}[t]{0.52\linewidth}
        \centering
        \includegraphics[width=\linewidth]{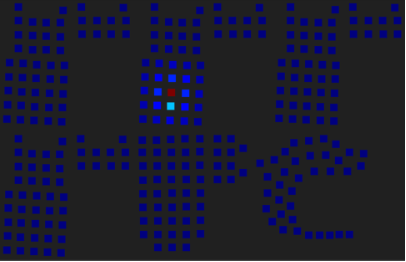}
        \caption{Simulated whole-hand taxels.}
        \label{fig:simulated_pressure_response}
    \end{subfigure}
    \hfill
    \begin{subfigure}[t]{0.46\linewidth}
        \centering
        \includegraphics[width=\linewidth]{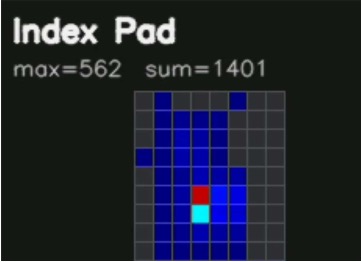}
        \caption{Real index-pad response.}
        \label{fig:real_pressure_response}
    \end{subfigure}
    \caption{Piezoresistive readouts: simulated whole-hand taxels and
    the physical index-pad response.}
    \label{fig:piezoresistive_response}
\end{figure}

Fig.~\ref{fig:piezoresistive_response} presents the pressure readouts.
The simulated view exposes the whole-hand taxel layout, while the real
readout focuses on the index pad and shows a localized response across
neighboring sensing elements. The two views differ in spatial extent and
display scale, so their colors are not a calibrated force comparison.

\subsection{Tactile Learning Examples}

Fig.~\ref{fig:result_b} shows in-hand ball rotation trained with depth
observations from all five fingertips. These observations describe the local
contact geometry at each fingertip as the hand manipulates the ball.
Fig.~\ref{fig:result_c} shows object grasping trained with full-palm tactile
sensing and fingertip depth. Palm pressure adds information about contacts
outside the fingertip cameras, complementing their local geometric detail.
Both examples use observations generated by the same simulator and
demonstrate its use for policy training in simulation. Policy-level
sim-to-real transfer remains a direction for future evaluation.

\section{Conclusion}

We presented RevoSim, a scalable simulator that jointly models the Revo3's
piezoresistive arrays and visuotactile fingertips. Its geometry-aware pressure
model and adaptive contact sampling provide consistent multimodal observations,
reduce depth VRAM by 70\% at 1,024 environments, and support depth-conditioned
marker-displacement observations in 4,096 parallel environments within
31.36~GiB. Future work will extend RevoSim to additional sensors, deformable
objects, and richer temporal dynamics.

\bibliographystyle{IEEEtran}
\bibliography{references}

\end{document}